%% file: main.tex
\documentclass[letterpaper, 10 pt, conference]{ieeeconf}

\IEEEoverridecommandlockouts                              
\usepackage{times}
\usepackage{multicol}
\usepackage{url}
\usepackage{color,soul}

\usepackage[dvipsnames]{xcolor}
\usepackage{subcaption}
\usepackage{amsmath,amssymb}

\usepackage{amsthm}
\usepackage{graphicx}  

\usepackage[colorinlistoftodos]{todonotes}
\newtheoremstyle{myplain}
    {}
	{}
	{\itshape}
	{}
	{\bfseries}
	{}
	{5pt plus 1pt minus 1pt}
	{}

\newtheoremstyle{mydefinition}
	{}
	{}
	{\normalfont}
	{}
	{\bfseries}
	{}
	{5pt plus 1pt minus 1pt}
	{}
	
\theoremstyle{myplain}
\newtheorem{definition}{Definition}
\newtheorem{theorem}{Theorem}

\let\labelindent\relax
\usepackage{enumitem}

\usepackage[bookmarks=true]{hyperref}

\input{notation}

\input{macros}

\title{\LARGE
Technical Report: A Hierarchical Deliberative-Reactive System Architecture for Task and Motion Planning in Partially Known Environments
}

\author{Vasileios Vasilopoulos$^{1}$, Sebastian Castro$^{1}$, William Vega-Brown$^{2}$, Daniel E. Koditschek$^{3}$, Nicholas Roy$^{1}$

\thanks{$^{1}$Computer Science and Artificial Intelligence Laboratory (CSAIL), MIT, Cambridge, MA 02139 \texttt{\{vvasilo,scastro, nickroy\}@csail.mit.edu}.}%
\thanks{$^{2}$Tagup, Inc. \texttt{will@tagup.io}.}%
\thanks{$^{3}$GRASP Laboratory, University of Pennsylvania, Philadelphia, PA 19104 \texttt{kod@seas.upenn.edu}.}%
\thanks{This work was supported by AFOSR grant FA9550-19-1-0265, the ARL/GDRS RCTA project, Coop. Agreement \#W911NF-10-2-0016, and the Toyota Research Institute Award LP-C000765-SR.}
}

\renewcommand{\baselinestretch}{.965}

\begin{document}

\maketitle
\thispagestyle{empty}
\pagestyle{empty}

\begin{abstract}
We describe a task and motion planning architecture for highly dynamic systems that combines a domain-independent sampling-based deliberative planning algorithm with a global reactive planner. We leverage the recent development of a reactive, vector field planner that provides guarantees of reachability to large regions of the environment even in the face of unknown or unforeseen obstacles. The reachability guarantees can be formalized using contracts that allow a deliberative planner to reason purely in terms of those contracts and synthesize a plan by choosing a sequence of reactive behaviors and their target configurations, without evaluating specific motion plans between targets. This reduces both the search depth at which plans will be found, and the number of samples required to ensure a plan exists, while crucially preserving correctness guarantees. The result is reduced computational cost of synthesizing plans, and increased robustness of generated plans to actuator noise, model misspecification, or unknown obstacles. Simulation studies show that our hierarchical planning and execution architecture can solve complex navigation and rearrangement tasks, even when faced with narrow passageways or incomplete world information.
\end{abstract}


\input{1-introduction-v2} 
\input{2-related-work}
\input{3-architecture-v3}

\input{4-implementation}

\input{5-results}

\input{6-conclusion}

\small
\bibliographystyle{IEEEtran}
\bibliography{references}


\end{document}

%% file: notation.tex
\newcommand{\Booleans}{\mathbb{B}}

\newcommand{\SETwo}{\mathrm{SE}(2)}

\newcommand{\CDRState}{\mathbf{s}}
\newcommand{\CDRStates}{\mathcal{S}}

\newcommand{\CDRBareAction}{a}
\newcommand{\CDRParams}{\boldsymbol{\theta}}
\newcommand{\CDRParamSpace}{\boldsymbol{\Theta}}

%% file: macros.tex
\newcommand{\robotradius}{r} 

\newcommand{\robotposition}{\mathbf{x}} 

\newcommand{\configuration}{\mathbf{c}} 

\newcommand{\robotorientation}{\psi} 

\newcommand{\goalposition}{\mathbf{x}^*} 

\newcommand{\goalconfiguration}{\mathbf{c}^*} 

\newcommand{\workspace}{\mathcal{W}} 
\newcommand{\freespace}{\mathcal{F}} 

\newcommand{\diffeogeneric}{\mathbf{h}} 

\newcommand{\controlfullyactuated}{\mathbf{u}} 
\newcommand{\controlunicycle}{\overline{\controlfullyactuated}} 
\newcommand{\controlfullyactuatedmodel}{\mathbf{v}} 
\newcommand{\controlunicyclemodel}{\overline{\controlfullyactuatedmodel}} 
\newcommand{\linearinput}{v} 
\newcommand{\angularinput}{\omega} 
\newcommand{\linearinputmodel}{\hat{\linearinput}} 
\newcommand{\angularinputmodel}{\hat{\angularinput}} 



%% file: 1-introduction-v2.tex
\section{Introduction}

\subsection{Motivation}

\begin{figure}
  \vspace{6pt}
  \centering
    \begin{subfigure}[t]{.45\columnwidth}
      \centering
      \includegraphics[width=\textwidth]{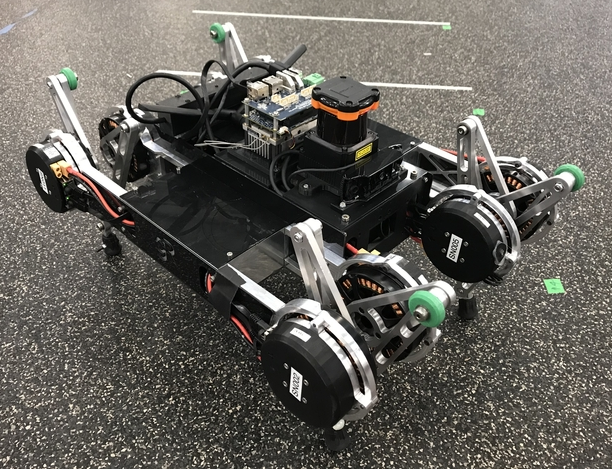}
      \caption{The Minitaur platform.}
      \label{fig:minitaur_real}
    \end{subfigure}
    \begin{subfigure}[t]{.44\columnwidth}
      \centering
      \includegraphics[width=\textwidth]{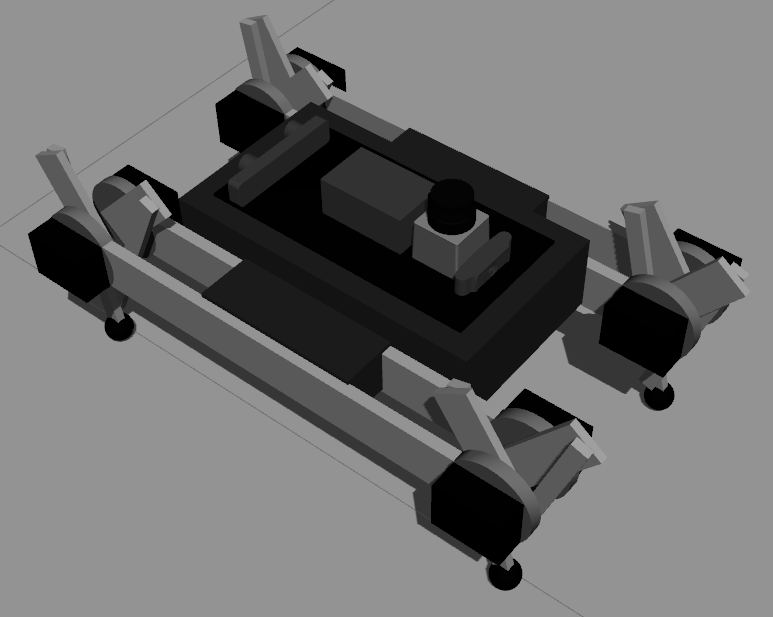}
      \caption{Simulated Minitaur.}
      \label{fig:minitaur_gazebo}
      \vspace{1pt}
    \end{subfigure}
    \begin{subfigure}[t]{.435\columnwidth}
      \centering
      \includegraphics[width=\textwidth]{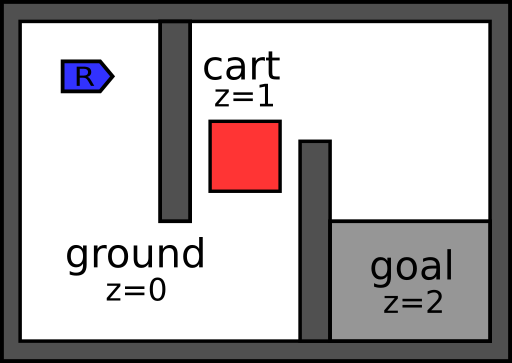}
      \caption{Simple planning problem.}
      \label{fig:toy_problem}
    \end{subfigure}
    \ 
    \begin{subfigure}[t]{.435\columnwidth}
      \centering
      \includegraphics[width=\textwidth]{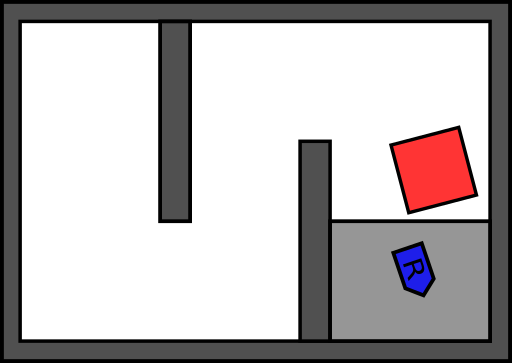}
      \caption{Possible goal state.}
      \label{fig:toy_problem_solution}
    \end{subfigure}
  \caption{%
    (a) The physical and (b) simulated Minitaur quadrupedal platform \cite{ghostminitaur}.
    (c) An example of a complex dynamically-constrained domain in which a robot must reach the goal on the lower right, where the heights of the ground plane ($z=0$), the obstacle ($z=1$), and the goal ($z=2$) are all different. The robot can jump only one unit, so the robot must move the cart (red square) to enable a sequence of jumps to reach a goal state (d).
    \label{fig:minitaur}
    \vspace{-18pt}
  }
\end{figure}

In this work, we consider a setting in which a highly energetic quadrupedal robot, capable of behaviors like walking, trotting and jumping, is assigned mobile manipulation tasks in an environment cluttered with fixed obstacles and movable objects (see Fig. \ref{fig:minitaur}). 
Solving these tasks requires planning and execution of dynamical pedipulation (nonprehensile manipulation of the environment using general purpose legs)~\cite{Mason_Lynch_1993} as well as navigation amidst clutter. 

Developing computationally and physically viable solutions for these scenarios is challenging, even assuming a deterministic robot in a fully observable world (e.g., PSPACE hardness of the Warehouseman's problem was established in \cite{Hopcroft_Schwartz_Sharir_1984}), and it has been well-understood for many years that hierarchical abstractions \cite{Wolfe2010} are required to address the fundamental complexity of such task and motion planning (TAMP) problems \cite{Kaelbling2011}. However, when using hierarchical abstract planners, it is difficult to ensure the correctness of the resulting plan unless the entire trajectory of the motion primitive is checked for feasibility during planning, significantly impacting the overall computational cost.

\begin{figure}[t]
\centering
\includegraphics[width=1.0\columnwidth]{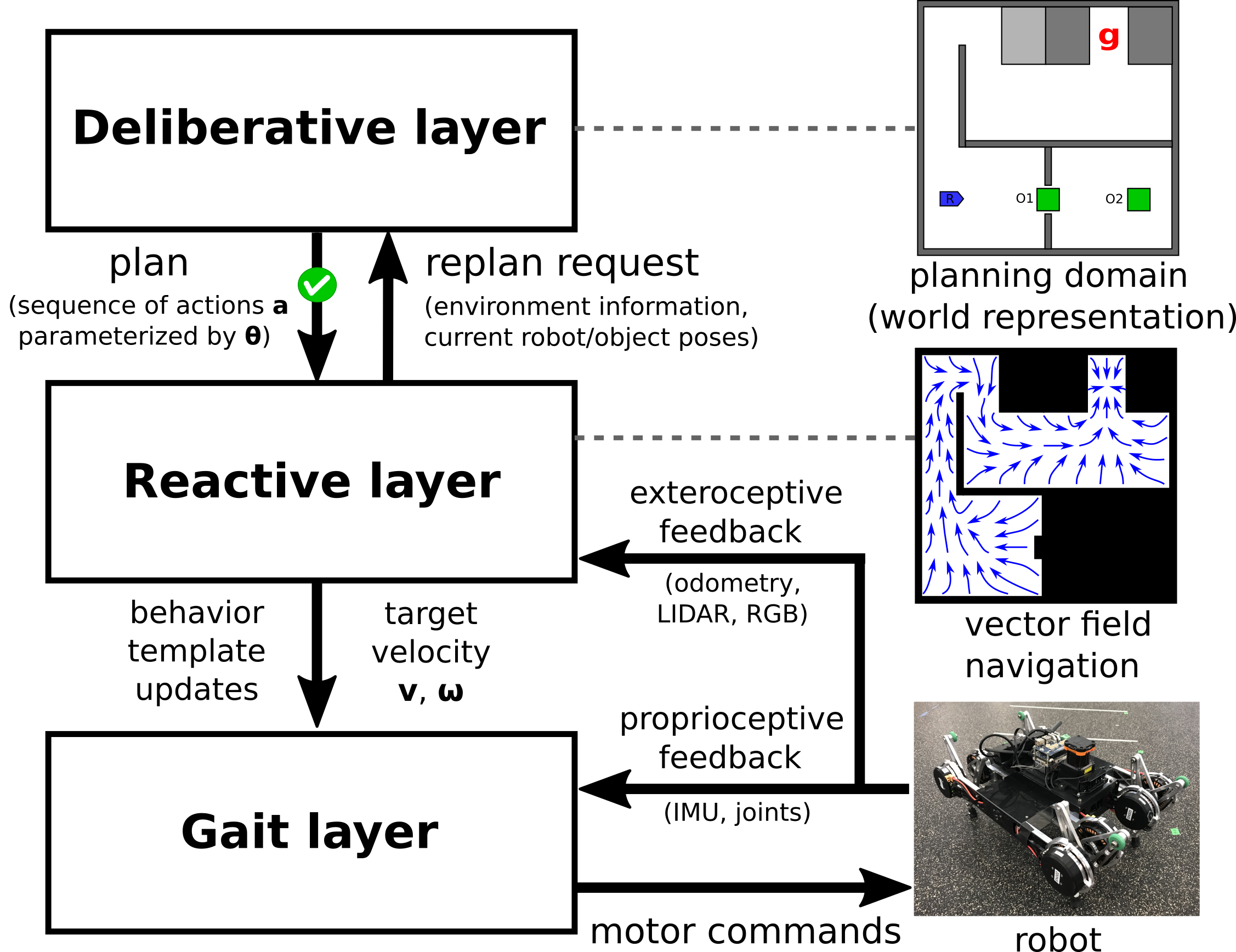}
\caption{\textbf{Our proposed system architecture}. 
Given a mobile manipulation task, a \emph{deliberative layer} searches for a sequence of abstract actions, or a \emph{plan}, using \emph{contracts} that describe the reachability guarantees of a global \emph{reactive layer}, which in turn implements these actions and guarantees collision avoidance in complex environments.
The reactive layer transmits template commands (such as target velocity or grasping commands) to a \emph{gait layer} that executes high-rate feedback to achieve parameterized steady-state or transitional behaviors on the robot. This architecture allows the deliberative layer to reason about sequencing actions without constructing explicit trajectories through the configuration space, improving computational efficiency while preserving probabilistic completeness.
}
\label{fig:architecture}
\vspace{-18pt}
\end{figure}

This paper shows how a deliberative layer and reactive layer can create abstract plans that are correct-by-construction through the use of continuous constraint contracts (C3) between a deliberative and reactive layer~\cite{vega-brown2020task}. For the reactive layer, we adapt a reactive, vector field planner from our prior work~\cite{vasilopoulos_pavlakos_bowman_caporale_daniilidis_pappas_koditschek_2020} that not only guarantees collision-free convergence to targets but is also robust to environmental uncertainty, even in the presence of unanticipated obstacles.

\subsection{Contributions}

The contribution of this paper is a hierarchical planner, shown in Fig.~\ref{fig:architecture}, that has the desirable properties of achieving the computational efficiency seen in many previous task and motion planning approaches, while preserving guarantees of probabilistic completeness that are often sacrificed for such computational gains. Our planner uses the guarantees of an online, vector-field-based reactive layer to define {\it action contracts}, such as the reachability of target poses, to an offline deliberative layer, providing the deliberative planner with global knowledge of each parameterized action's basin of attraction, and allowing it to simply reason about sequencing actions without constructing explicit trajectories through the configuration space.

Focusing on the example of a quadruped robot navigating in an environment with static and movable obstacles, we demonstrate several significant computational advantages in the deliberative layer arising from delegating metric details to the reactive (closed loop) controller.
First, the reactive layer allows the deliberative planner to plan only in terms of transitions between behaviors, such as grasping and releasing objects. As a result, fewer samples are needed 
to find a good plan. 
Second, because the deliberative planner is aware of the domain of convergence for each controller and can work out plans using the adjacency of those basins of reachability, it can construct plans with fewer steps than if it relied solely on fine-grained motion primitives---for example, line-of-sight connections---and without sacrificing any correctness guarantees. The difference is especially pronounced when the path requires traversing narrow passages, which are notoriously difficult for sampling-based planners. 
The reduction in the length of plans dramatically reduces the time required to search for a high-level plan. 

\subsection{Organization of the Paper}

The paper is organized as follows. Section~\ref{sec:related-work} summarizes related work. Section~\ref{sec:architecture} describes the proposed multi-layer architecture, along with its formal guarantees. Section~\ref{sec:domain_description} describes the mobile manipulation problems addressed in this paper, and the specific set of introduced symbolic actions for the Ghost Minitaur robot~\cite{ghostminitaur} used in this study, in two different planning approaches for the deliberative planner: a {\it local reactive} approach, employing the reactive layer to simply track reference trajectories from the deliberative layer, and our {\it global reactive} approach with the proposed architecture. Section~\ref{sec:numerical_experiments} describes numerical studies contrasting the performance of the local reactive planner with the global reactive planner in different mobile manipulation scenarios. Section~\ref{sec:physical_experiments} provides implementation examples with a simulated Minitaur, 
and, finally, Section~\ref{sec:conclusions} concludes with our remarks and ideas for future research.

%% file: 2-related-work.tex
\section{Related Work}
\label{sec:related-work}

Hierarchical abstractions for TAMP have been well studied in the literature.
Examples include the use of a deliberative planner that employs a reactive execution layer---such as a motion primitive library \cite{Majumdar_Tedrake_2017}, or pre-image backchaining with a higher-level planner in deterministic \cite{Kaelbling2011} and stochastic \cite{kaelbling2017pre} settings---to simplify the computational burden of planning. 
Solutions typically involve a marriage of fast discrete planning tools~\cite{dantam2018incremental} and sampling- or grid-based discretizations of the continuous action space~\cite{garrett2020pddlstream}, with significant engineering effort expended on the design of effective heuristics and sampling strategies that exploit task-level and geometric information~\cite{lagriffoul2016combining}. Angelic semantics \cite{marthi2008} provide a way of describing abstractions that also preserve optimality, but there is no easy way of defining such abstractions in continuous domains. Our prior work \cite{vega-brown2016asymptotically} provided a step towards tractable planning with complex kinematic constraints, but no appropriate approach exists for the complex legged robot dynamics considered in this paper.

Motivated by the typically high-dimensional configuration spaces arising from combined task and motion planning, most approaches focus either on sampling-based methods that empirically work well \cite{vandenBerg2010,Krontiris-RSS-15}, or learning a symbolic language on the fly \cite{konidaris2018}. Such methods require constant replanning in the presence of unanticipated conditions and their search time grows exponentially with the number of configuration variables.

Other approaches focus on the use of reactive temporal logic planning algorithms \cite{lahijanian2016iterative,livingston2012backtracking,kress-gazit-2009}, that can account for environmental uncertainty in terms of incomplete environment models, and also ensure correctness when the robot operates in an environment that satisfies the assumptions modeled in the task specification. Common in these works is the reliance on discrete abstractions of the robot dynamics  \cite{belta2005discrete,pola2008approximately}, while active interaction with the environment to satisfy the logic specification is neglected.

%% file: 3-architecture-v3.tex
\section{Vector-Field Task Planning}
\label{sec:architecture}

Our objective is to compute plans for a robot to achieve a goal state, subject to kinodynamic constraints. The active constraints on the dynamics of the world state vary with the robot's behavior, enabling the robot to select different {\it modes} of its dynamics as it plans to move around the world. For example, a plan for the robot in Fig.~\ref{fig:minitaur} might simply be to navigate its workspace, or to make and break contact with the objects in the world as it moves around. Each of these modes corresponds to a different set of constraints.

Following the notation introduced in our prior work \cite{vega-brown2016asymptotically}, we can define a {\it planning domain} by a tuple $(h,\mathcal{C})$, where $h:\mathcal{C} \times T\mathcal{C} \rightarrow \mathbb{R}^k$ defines a set of $k$ constraints on the configuration space $\mathcal{C}$ and its tangent bundle $T\mathcal{C}$. Then, a differentiable function $\sigma : [0,T] \rightarrow \mathcal{C}$ is a {\it feasible path} if $h(\sigma(t),\dot{\sigma}(t)) \geq 0, \forall t \in [0,T]$, where $\dot{\sigma}(t) = d\sigma(t) / d t$. 
We denote the set of feasible paths by $\Sigma_\mathcal{C}$.

Based on this description, we can define a {\it planning problem} as a tuple $(\mathbf{c}_0, \mathbf{c}^*)$, where $\mathbf{c}_0, \mathbf{c}^* \in \mathcal{C}$ are the initial and goal configurations respectively. 

\begin{definition}
\label{definition:planning_problem}
A solution to a planning problem $(\mathbf{c}_0, \mathbf{c}^*)$ in a domain $(h,\mathcal{C})$ is given by the solution to the problem:
\begin{align}
    \mathrm{find} \quad & \sigma \in \Sigma_\mathcal{C} \label{eq:optimization_problem} \\
    \mathrm{s.t.} \quad & h(\sigma(\tau),\dot{\sigma}(\tau)) \geq 0, \quad \forall \tau \in [0,T] \nonumber \\
    & \sigma(0) = \mathbf{c}_0, \quad \sigma(T) = \mathbf{c}^* \nonumber
\end{align}
\end{definition}

Problem~\eqref{eq:optimization_problem} is formally undecidable without further assumptions on $h(\sigma(\tau),\dot{\sigma}(\tau))$ \cite{vega-brown2020task}, and solving the analogous problem for typical discrete approximations is computationally intractable for scenarios where the robot needs to make and break contact with the environment. A conventional approach is to decompose the problem into a task and motion planning problem: a deliberative layer first solves for a task plan corresponding to a sequence of dynamic modes, parameterized by starting and stopping conditions, and a motion planner generates trajectories within each mode from start condition to stopping condition. 

For instance, for our quadrupedal robot manipulating objects as shown in Fig.~\ref{fig:minitaur}, we can solve  rearrangement or navigation problems as a sequence of traversals, jumping motions, and object grasps and movements. The deliberative layer determines the sequence of objects to interact with, and where to navigate to make or break contact with objects. The motion planner computes the specific trajectories the robot should follow to realize the sequence of mode changes.

However, the decomposition into separate task and motion planning problems typically leads to loss of completeness, because the task planner may create motion planning sub-problems that are infeasible. We now describe the formal conditions under which a combined deliberative and motion planning layer can compute task plans that preserve probabilistic completeness guarantees of the underlying motion planner, even without first evaluating it.

\subsection{Deliberative layer}
\label{sec:deliberative_layer}

We assume that the deliberative planner has knowledge of the entire configuration space, including a description of the world as a collection of objects with geometric information, such as shape and pose, and other properties that constrain the types of actions available with these objects.

We use the {\it continuous constraint contract} (C3) to represent states, presented in our prior work~\cite{vega-brown2020task}. The C3 representation is a continuous extension of the SAS+ formalism \cite{backstrom1995complexity}; as in most planning formalisms, the state of the world is parameterized by the \emph{value} of different \emph{variables}. A state $\CDRState \in \CDRStates$ is a collection of variable--value pairs, and represents the set of configurations satisfying the constraint defined by the value assigned to each variable. Each variable $v$ corresponds to a function $\eta_v(\mathbf{c})$ mapping configurations $\mathbf{c}$ to an element of the variable's domain; a state $\{v_1 = p_1, v_2 = p_2\}$ describes the set of configurations $\mathbf{c}$ such that $\eta_{v_1}(\mathbf{c}) = p_1$ and $\eta_{v_2}(\mathbf{c}) = p_2$.  For instance, for our example in Fig.~\ref{fig:minitaur}, if a specific object is given a specific pose in the environment, then the state will be all configurations of the world (including all configurations of the robot) that have that object at that pose. There is no requirement that every variable have an assigned value; variables without values represent inactive constraints. This implies that a single configuration $\mathbf{c}$ could satisfy the requirements of multiple states at once; e.g., a configuration $\mathbf{c}$ that satisfies $\eta_{v_1}(\mathbf{c}) = p_1$ and $\eta_{v_2}(\mathbf{c}) = p_2$ corresponds to both states $\{v_1 = p_1, v_2 = p_2\}$ and $\{v_1 = p_1\}$. However, the latter state describes a larger set of configurations, with the constraint on $v_2$ being inactive.

To travel between states, we assume the lower level motion planner can instantiate \mbox{{\it actions}}, parameterized by a start and goal, in the form of different \mbox{{\it behaviors}}, such as \mbox{{\it ``Navigate''}} or \mbox{{\it ``Move-Object''}}; the deliberative planner must then choose a sequence of different behaviors as well as their parameterization. This representation is equivalent to the representation used in Definition \mbox{\ref{definition:planning_problem}} for controllable systems.
\begin{theorem}[Representing constraints as analytic functions -- Included in \cite{vega-brown2020task}]
\label{theorem:representation}
If the kinodynamic constraints $h$ are piecewise-analytic in the sense of Sussmann \cite{sussmann_1979}, and the dynamical system is stratified controllable in the sense of Goodwine and Burdick \cite{goodwine_burdick_2001}, then there is a stratified C3 instance whose actions represent piecewise-analytic vector fields, in which the constraints can be expressed as equalities and inequalities involving only analytic functions.
\end{theorem}

The consequence of this theorem is that if the constraints are piecewise-analytic, and if the dynamics are stratified, then the complete plan can be expressed as a series of motions across a sequence of vector fields, where the vector fields are defined by the kinodynamic constraints such as obstacles or grasping and releasing objects. The planning problem then becomes one of choosing the sequence of vector fields and their parameterizations.

We can further take advantage of this theorem by defining actions as a {\it contract} between the deliberative and motion layers: formally, we define the requirements and effects of an action $\CDRBareAction$ with continuous parameterization $\CDRParamSpace_{\CDRBareAction}$ in terms of two functions $g_{\CDRBareAction}: \CDRStates \times \CDRParamSpace_{\CDRBareAction} \to \Booleans$ and $f_{\CDRBareAction}: \CDRStates \times \CDRParamSpace_{\CDRBareAction} \to \CDRStates $, where $\CDRStates$ is the space of possible states $\CDRState$. The function $g_{\CDRBareAction}(\CDRState, \CDRParams)$ defines the \emph{requirements} of the action, and $f_{\CDRBareAction}(\CDRState, \CDRParams)$ defines its \emph{effects}.  If the system is in state $\CDRState$ when executing action $\CDRBareAction$ with parameters $\CDRParams$, then the motion planner guarantees that if $g_{\CDRBareAction}(\CDRState, \CDRParams) = 1$ then at some point in the future the system will reach state $\CDRState' = f_{\CDRBareAction}(\CDRState, \CDRParams)$.

However, such guarantees are in practice difficult to describe. 
The easiest guarantee to provide is one where the motion planner is restricted to straight-line actions parameterized by an end point, and enforces reachability by evaluating each straight-line trajectory for violations of the kinodynamic constraints. 
A deliberative planner using this simplistic motion planner would offer probabilistic guarantees, but with essentially no computational advantage from the decomposition into deliberative and motion planning. 
The challenge is to identify a motion planner that can enforce the C3 contracts in a computationally efficient manner.

\subsection{Reactive layer}
\label{sec:reactive_layer}

We now describe a reactive motion planner with a key property: the corresponding C3 contracts can be checked very quickly, without sampling, discretization, or collision-checking. Rather than instantiating a single motion plan, the reactive layer constructs a control policy whose execution is guaranteed to achieve the objectives specified by the deliberative layer, or to return with a failure condition expressing the incorrectness of a presumed constraint in the actual environment. The reactive planner can reliably handle the geometric details of navigation and manipulation, even in the face of unknown obstacles. 

The reactive layer models the robot as a polygon, and takes as input an estimate of the current reachable set of robot poses, in the form of a polygonal connected component of the robot's workspace, along with a high-level action $\CDRBareAction$ with all parameters $\CDRParams \in \CDRParamSpace_{\CDRBareAction}$ chosen by the deliberative layer. Following our example from Fig.~\ref{fig:minitaur}, one action in a plan for a rearrangement task might be $\texttt{push}(\texttt{ground}, \texttt{cart}, \robotposition^*)$ (interpreted as ``push object \texttt{cart} to a target location $\robotposition^*$ atop the object \texttt{ground}''). The reactive layer outputs the parameters of a behavior, such as a target velocity and yaw rate, which vary continuously with the state of the world. 

The reactive layer is implemented using the vector-field-based feedback motion planning scheme introduced in our prior work~\cite{vasilopoulos_pavlakos_bowman_caporale_daniilidis_pappas_koditschek_2020}. Its critical advantage is the use of a diffeomorphism construction to deform non-convex environments to easily navigable convex worlds, by employing a ``crude'' geometric description of the environment (outer walls) from the deliberative layer, together with learned or intrinsic domain specific knowledge about encountered obstacles. 

Unlike the deliberative layer's more general world representation including the robot and all movable objects and static obstacles, in the reactive layer we assume that the robot is the only active agent in the world, and behaves like a first-order, nonholonomically-constrained, disk-shaped robot, centered at location $\robotposition \in \mathbb{R}^2$, with radius $\robotradius \in \mathbb{R}_{>0}$, orientation $\robotorientation \in S^1$ and input vector $\controlunicycle:=(\linearinput,\angularinput)$, consisting of a fore-aft and an angular velocity command. 
We denote by $\workspace$ the robot's non-convex polygonal workspace, and by $\workspace_{\robotposition} \subseteq \workspace$ the polygonal region corresponding to the space reachable from the robot's current position $\robotposition$. The workspace is cluttered by a finite collection of disjoint obstacles of unknown shape, number, and placement. Similarly, the \textit{freespace} $\freespace$ is defined as the set of collision-free placements for the robot in $\workspace$, and we denote by $\freespace_{\robotposition} \subseteq \freespace$ the freespace component corresponding to $\workspace_{\robotposition}$.

During online execution, the reactive controller synthesizes an action as a control law by constructing a diffeomorphism $\diffeogeneric$ between $\freespace_{\robotposition}$ and a convex \emph{model environment}, where non-convex obstacles are either deformed to topologically equivalent disks or merged to the boundary of $\freespace_{\robotposition}$. Then, the robot can navigate by generating virtual commands $\controlunicyclemodel := (\linearinputmodel,\angularinputmodel)$ as in \cite{Arslan_Koditschek_2018}, for an equivalent unicycle model (defined in \cite[Eqs. (24)-(25)]{vasilopoulos_koditschek_2018}) that navigates toward the assigned target position $\goalposition$ in this model environment, and then mapping the virtual commands to physical inputs $(\linearinput,\angularinput)$ through the push-forward of the inverse of $\diffeogeneric$, i.e., $\controlunicycle = \left[ D_{\robotposition} \diffeogeneric \right]^{-1} \controlunicyclemodel$. 

Using the language of the deliberative layer, the requirements $g_a(\robotposition,\goalposition)$ of a navigation action are satisfied if both the robot and target positions are contained in the same component $\freespace_{\robotposition}$ of the robot's freespace. Because we represent the workspace and obstacles using polygons, this calculation can be made using standard tools from computational geometry. More formally, we can decompose $\freespace$ into a finite collection of connected polygons (possibly with holes), and define a set-valued function $\beta_{\freespace}: \freespace \to 2^\freespace$, such that $\freespace_{\robotposition} \triangleq \beta_\freespace(\robotposition) \subseteq \freespace$ is the connected component containing $\robotposition$. We describe an implementation of this function in Section~\ref{sec:global_reactive_planning}.

\begin{theorem}[Target convergence and obstacle avoidance -- Corollary of \mbox{\cite[Theorem 2]{vasilopoulos_pavlakos_bowman_caporale_daniilidis_pappas_koditschek_2020}}]
\label{theorem:convergence}
If we define $g_a(\robotposition,\goalposition)$ to be equal to 1 when $d(\beta_{\freespace}(\robotposition), \beta_{\freespace}(\goalposition)) = 0$ and 0 otherwise (with $d(\cdot,\cdot)$ the distance between sets), then the online reactive planner guarantees that the robot will converge to the target $\goalposition$ (i.e., $f_a(\robotposition) \triangleq \goalposition$), while avoiding all obstacles in its workspace.
\end{theorem}

In other words, during online execution, as long as both the initial robot position $\robotposition$ and the target location $\goalposition$ lie in the same polygonal component $\freespace_{\robotposition}$ of the robot's freespace, the robot can construct a diffeomorphism to the environment that allows it to compute a controller that will eventually reach $\goalposition$ while avoiding all (previously known or unforeseen) obstacles in the environment. This result also holds for rearrangement actions, using the framework in \cite{Vasilopoulos_Topping_Vega-Brown_Roy_Koditschek_2018} for generating virtual commands for the center of the circumscribed disk, enclosing the robot and a grasped disk-shaped object.

It should be noted that Theorem~\ref{theorem:convergence} covers only navigation actions, where either the robot or the robot-object pair navigates to a desired location. To navigate across mode boundaries (i.e., across connected components of the configuration space), we use special-purpose local actions. For example, continuing to follow Fig.~\ref{fig:minitaur}, we assume the existence of an action that can take the robot from \texttt{ground} to \texttt{cart}. Unlike for navigation, these local planners are not formally verified in the sense of Theorem~\ref{theorem:convergence}, but may similarly be designed to use robust closed-loop mechanisms during real-world execution. The reader is referred to Section~\ref{sec:domain_description} for a description of such an action (\texttt{jump}).

\subsection{Combined deliberative-reactive planning}
\label{sec:combined_search}

In practice, given a C3 problem instance describing the permitted behaviors, the world geometry, and a goal specification, we construct a graph by sampling random parameters for actions.
In navigation or manipulation problems, this generally involves sampling candidate placements for objects or for the robot. Importantly, Theorem~\ref{theorem:representation} allows us to sample from the free space of each object independently, rather than sampling from the joint configuration space, without sacrificing completeness.
The sampling process implicitly defines a planning graph: by construction, the finite set of sampled parameters yields a finite (though exponentially large) set of potentially reachable states.
Each distinct state is a node in the graph, connected with an edge to another state that satisfies the requirements of the action provided by the reactive planner.
We then perform a direct heuristic search over this graph to synthesize a plan.
We reduce computational cost by considering only a reduced set of constraints in the action requirements when sampling, and checking the remaining constraints only when we find a candidate plan to a given state. 
Although the graph grows exponentially with number of samples, we typically need to construct and search only a small fraction of the graph. 
We refer the reader to \cite{vega-brown2016asymptotically} for more details on the graph construction and search.

\begin{theorem}[Combined probabilistic completeness -- Corollary of Theorems~\ref{theorem:representation}--\ref{theorem:convergence}] \label{theorem:completeness}
If our planning domain contains only modes defined by piecewise-analytic constraints and stratified controllable dynamics, and there exist local actions for navigating across mode boundaries, then the deliberative planner will eventually sample a feasible motion plan, expressed as a sequence of reactive planner actions between connected components of the configuration space.
\end{theorem}

Combining the deliberative and reactive layers within this sampling-based framework yields a planning algorithm that is provably sound, probabilistically complete, and synthesizes feedback control policies robust to perturbation from the environment and actuator noise.

%% file: 4-implementation.tex
\section{System Implementation}
\label{sec:domain_description}
In this Section, we describe the specific class of mobile manipulation problems addressed in this work, and the implementation details of the above architecture. While our planner is general purpose for a wide range of problems, 
we consider the problem of a dynamically complex quadruped robot performing navigation among movable obstacles (NAMO) as in our prior work \cite{Vasilopoulos_Topping_Vega-Brown_Roy_Koditschek_2018}.

\subsection{Problem Domain Description}
\label{sec:problem_description}

Our chosen model abstraction for planning is a 2.5D \emph{semiplanar world} representation, shown in Fig.~\ref{fig:toy_problem}.
All objects in the world, which can be either static or movable, are defined as planar polygons, with a pose in $\SETwo$ augmented by a $z$ value denoting vertical height.  
The robot can walk along the polygonal component describing the top of the currently occupied object, jump on and off the ground plane and between objects of varying heights provided the height difference and gap is within its physical capabilities, and manipulate movable objects on the currently occupied object.
We use five types of actions to model the robot capabilities:
    
    \textbf{Moving:} For each object \texttt{obj} in the world that the robot can occupy, we include an action \texttt{move(base, obj, p\_target)}, which requires feasibility of motion from the current pose to a new pose \texttt{p\_target}. 
    Formally, this action guarantees transition from a state $\CDRState = \{ \texttt{atop}(\texttt{pose}(\texttt{obj}),\texttt{base}) = 1 \}$ to a new state $\CDRState' = \{ \texttt{atop}(\texttt{pose}(\texttt{obj}),\texttt{base}) = 1$, $\texttt{pose}(\texttt{obj}) = \texttt{p\_target} \}$, given the requirement $\texttt{isfeasible}(\texttt{pose}(\texttt{obj}),\texttt{p\_target})=1$. The requirement \texttt{isfeasible} can either denote feasibility in the typical sense of motion planning that involves sampling-based collision checking (as explained in Section~\ref{sec:local_reactive_planning}), or feasibility in the sense of our reactive planner (as explained in Section~\ref{sec:reactive_layer} and later in Section~\ref{sec:global_reactive_planning}).
    
    \textbf{Jumping:} An action \texttt{jump(from, to, p\_target)} requires that the robot can feasibly jump from object \texttt{from} to object \texttt{to} at pose \texttt{p\_target}. Formally, this action guarantees transition from a state $\CDRState = \{ \texttt{atop}(\texttt{pose}(R),\texttt{from}) = 1 \}$ to a new state $\CDRState' = \{ \texttt{pose}(R) = \texttt{p\_target},  \texttt{atop}(\texttt{pose}(R),\texttt{to}) = 1$ \}, given the requirement $||\texttt{pose}(R)-\texttt{p\_target} || \leq \epsilon$.
    
    \textbf{Grasping:} An action \texttt{grasp(base, obj, p\_target, p\_mount)} defines the mounting of a movable object \texttt{obj} on top of object \texttt{base} to a new pose \texttt{p\_target}. An additional parameter \texttt{p\_mount} defines the relative pose of the robot and the grasped object during motion. 
    Formally, this action guarantees transition from a state $\CDRState$, where both the robot $R$ and the object \texttt{obj} are on top of object \texttt{base} and the value of $\texttt{grasping}(R,\texttt{obj})$ is 0, to a new state $\CDRState'$, where $R$ and \texttt{obj} remain on top of \texttt{base}, $\texttt{grasping}(R,\texttt{obj}) = 1$, $\texttt{pose}(\texttt{obj}) = \texttt{p\_target}$ and $\texttt{relative\_pose}(R,\texttt{obj}) = \texttt{p\_mount}$. The requirement of this action is given by $\texttt{isfeasible}(\texttt{pose}(R),\texttt{pose}(\texttt{obj}))=1$, using the feasibility function described above.
    
    \textbf{Releasing:} An action \texttt{release(base, obj, p\_robot, p\_obj)} defines the releasing of a grasped object \texttt{obj} on top of object \texttt{base}, such that the robot ends in pose \texttt{p\_robot} and the object in pose \texttt{p\_obj}. Formally, this action guarantees transition from a state $\CDRState$, where both the robot $R$ and the object \texttt{obj} are on top of object \texttt{base} and $\texttt{grasping}(R,\texttt{obj}) = 1$, to a new state $\CDRState'$, where $R$ and \texttt{obj} remain on top of \texttt{base}, $\texttt{grasping}(R,\texttt{obj}) = 0$, $\texttt{pose}(\texttt{obj}) = \texttt{p\_obj}$, and $\texttt{pose}(R) = \texttt{p\_robot}$. The requirement of this action is given by $\texttt{isfeasible}(\texttt{pose}(\texttt{obj}),\texttt{\texttt{p\_obj}})=1$, using the same feasibility function.
    
    \textbf{Pushing:} Finally, we define an action \texttt{push(base, obj, p\_target)} in which a robot moves atop an object \texttt{base} with a grasped object \texttt{obj} towards a pose \texttt{p\_target}, without releasing the object. The requirements and effects of \texttt{push} are similar to those of \texttt{release}, without defining a desired final pose for the robot $R$.

To demonstrate our proposed architecture, we attempt to solve mobile manipulation problems where the reactive layer is instantiated with either a {\it local reactive approach} that does not allow the deliberative planner to query the reactive planner for motion contracts, or with the proposed {\it global reactive approach} described in Section~\ref{sec:architecture} (see Fig.~\ref{fig:reactive_planning_comparison}). We use the local reactive approach as a baseline comparison because it is equivalent to state-of-the-art TAMP planners that provide correctness guarantees, for example, \mbox{\cite{garrett2020pddlstream,vega-brown2016asymptotically,hauser2010multi}}.

\subsection{Local Reactive Planning Approach}
\label{sec:local_reactive_planning}

\begin{figure}
  \vspace{5pt}
  \centering
    \begin{subfigure}[t]{\columnwidth}
        \centering
        \includegraphics[width=0.875\columnwidth]{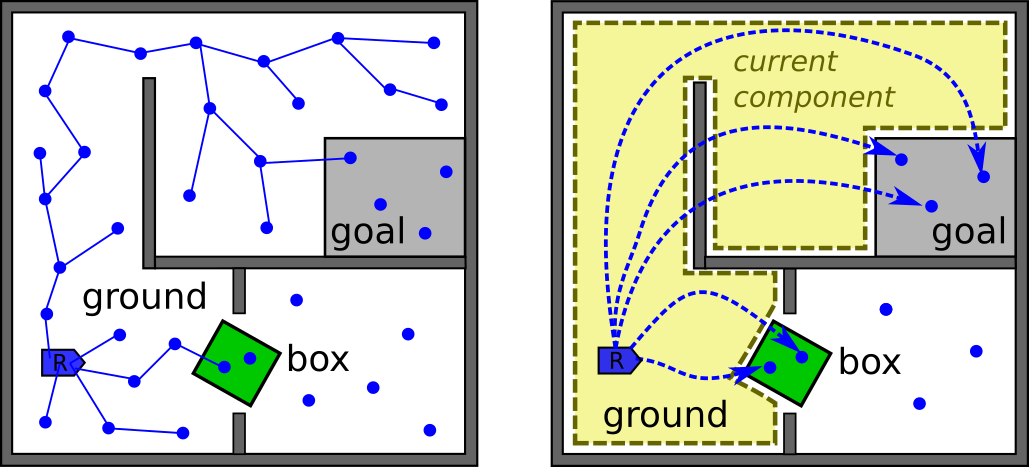}
        \caption{
        In our baseline local reactive planning approach (left), the deliberative planner must conduct an optimized search over the configuration space of robot and object placements in the presumed freespace and is restricted to  collision free straight-line paths.
        Motions that instantiate these paths are generated at runtime by the reactive layer, guaranteeing avoidance of unanticipated obstacles along the way. 
        The global reactive planning approach (right) is guaranteed to generate a collision-free path to any target pose in the robot's current connected component (highlighted in yellow). 
        Actions are now represented by putative robot-connected components and their adjacency relative to robot--object manipulations. This more abstract contract between layers reduces the deliberative planner's computational burden to the exploration of topological adjacency.}
        \label{fig:sampling_comparison}
        \vspace{2pt}
    \end{subfigure}
    \begin{subfigure}[t]{\columnwidth}
        \centering
        \includegraphics[width=\textwidth]{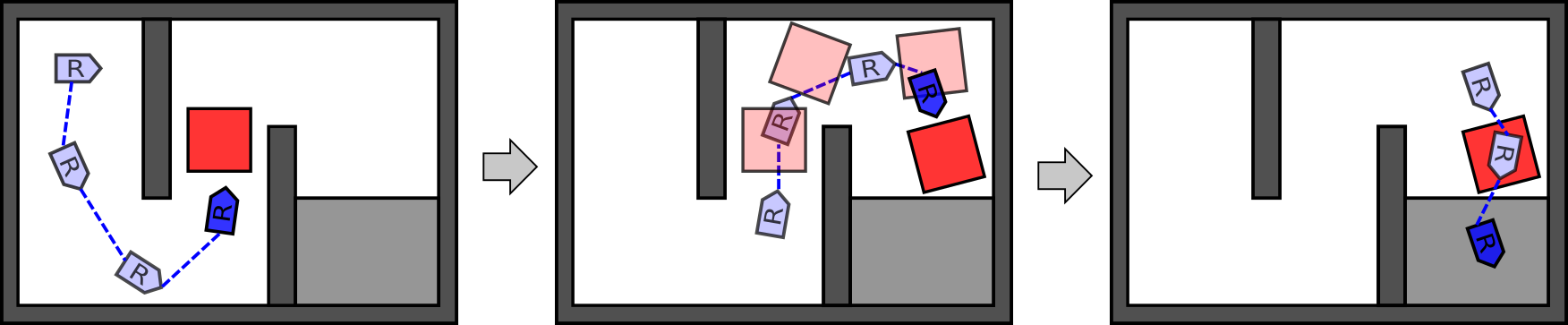}
        \caption{
        Solution to the example from Fig.~\ref{fig:minitaur} using the local reactive planning architecture (Section \ref{sec:local_reactive_planning}) \cite{Vasilopoulos_Topping_Vega-Brown_Roy_Koditschek_2018}. The deliberative planner finds a sequence of collision-free straight-line motion primitives to move the robot to the cart, push the cart near the goal, and finally jump to the goal.
        Resulting plans are often long sequences comprising the entire set of actions listed in Section~\ref{sec:problem_description}.}
        \label{fig:toy_example_integrated}
        \vspace{2pt}
    \end{subfigure}
    \begin{subfigure}[t]{\columnwidth}
        \centering
        \includegraphics[width=\textwidth]{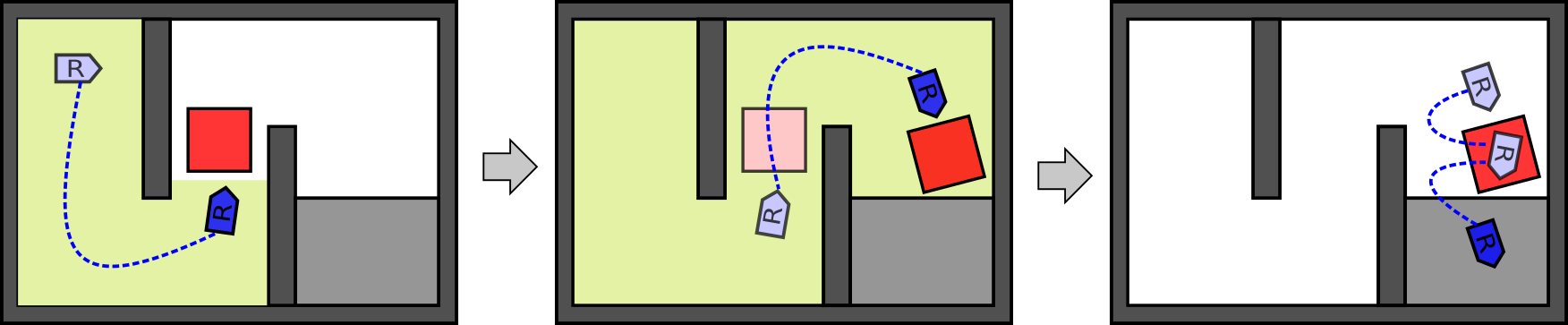}
        \caption{
        Solution to the example from Fig.~\ref{fig:minitaur} using the global reactive planning approach (Section \ref{sec:global_reactive_planning}).
        Shaded regions indicate the robot's currently occupied connected component, defining the (global) navigation domain for the reactive layer.
        The dashed lines are purely illustrative, as the actual paths are unknown to the deliberative planner and commanded at runtime by the reactive layer.
        }
        \label{fig:toy_example_hierarchical}
    \end{subfigure}
    \caption{Comparison of local and global reactive planning approaches.}
    \label{fig:reactive_planning_comparison}
    \vspace{-15pt}
\end{figure}

Given a sequence of actions from the deliberative layer, the local reactive planner only guarantees the feasibility of navigating from a starting pose to a target pose if the path between them is collision-free~(Fig.~\ref{fig:toy_example_integrated}). That is, the reactive planner can move from $\robotposition$ to $\goalposition$ if $\texttt{isfeasible}(\robotposition,\goalposition)= 1$, with $\texttt{isfeasible}(\robotposition_1,\robotposition_2)$ equal to 1 when $C_{\freespace,r}(P_{sweep}(\robotposition_1, \robotposition_2), \configuration) \geq 0$ and 0 otherwise, where $P_{sweep}(\robotposition_1,\robotposition_2)$ is a polygon containing the robot polygon at each pose along a geodesic between poses $\robotposition_1$ and $\robotposition_2$ in $\SETwo$, and $C_{\freespace,r}(P, \configuration)$ checks for collision of a polygon $P$ with any object or obstacle in configuration $\configuration$.
Here, the local reactive layer only plays its intermediating role when recovering the target pose in the face of unanticipated obstacles---or reporting the infeasibility of doing so. 

\begin{figure*}[ht]
  \vspace{2pt}
  \centering
    \begin{subfigure}[t]{0.1925\textwidth}
      \centering
      \includegraphics[width=\textwidth]{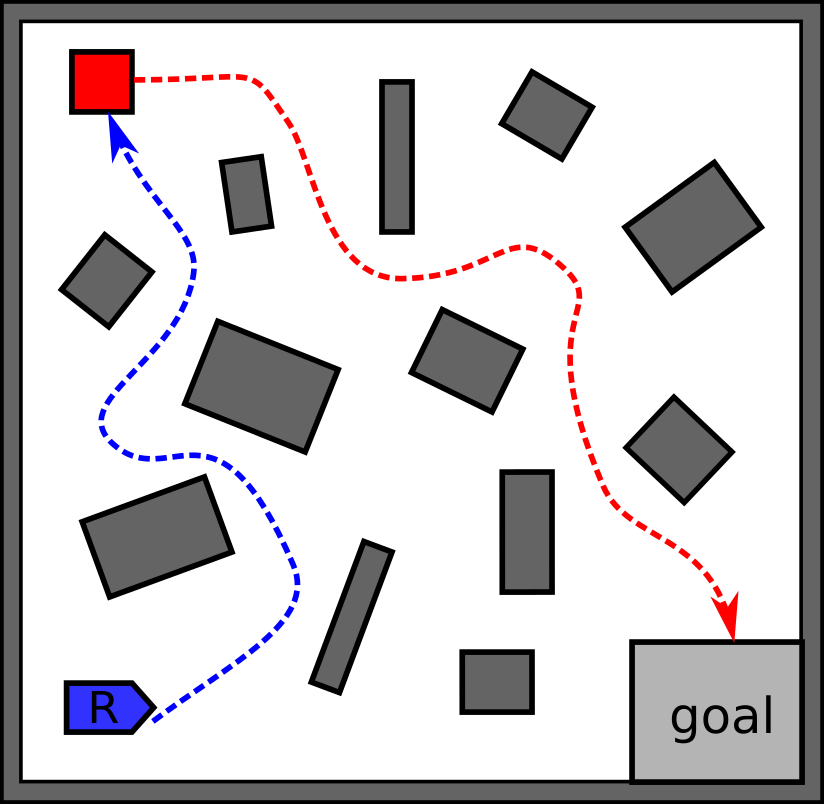}
        \caption{\small{Scenario overview}}
        \label{fig:scenario_obstacles}
    \end{subfigure}
    \begin{subfigure}[t]{0.263\textwidth}
        \includegraphics[width=\textwidth]{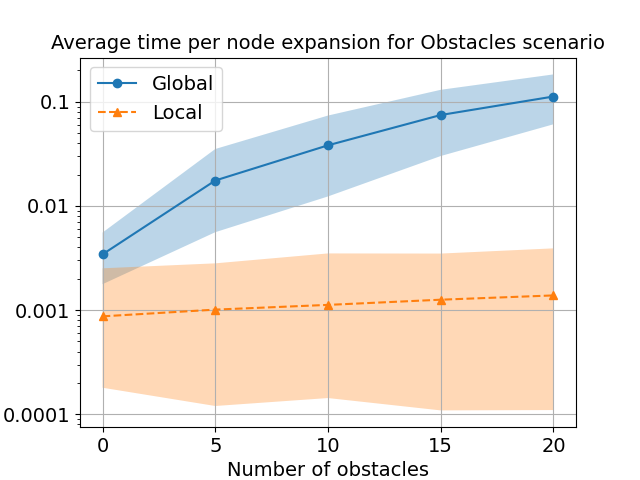}
        \caption{\small{Expansion time vs. complexity}}
        \label{fig:results_obstacles_node_time}
    \end{subfigure}
    \begin{subfigure}[t]{0.263\textwidth}
        \includegraphics[width=\textwidth]{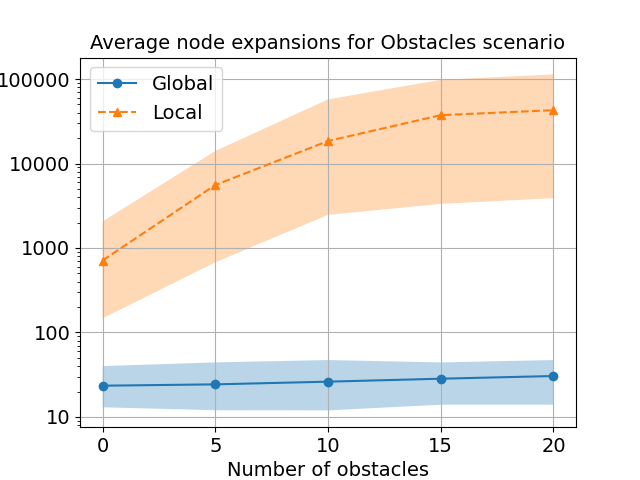}
        \caption{\small{Node expansions vs. complexity}}
        \label{fig:results_obstacles_node}
    \end{subfigure}
    \begin{subfigure}[t]{0.263\textwidth}
        \includegraphics[width=\textwidth]{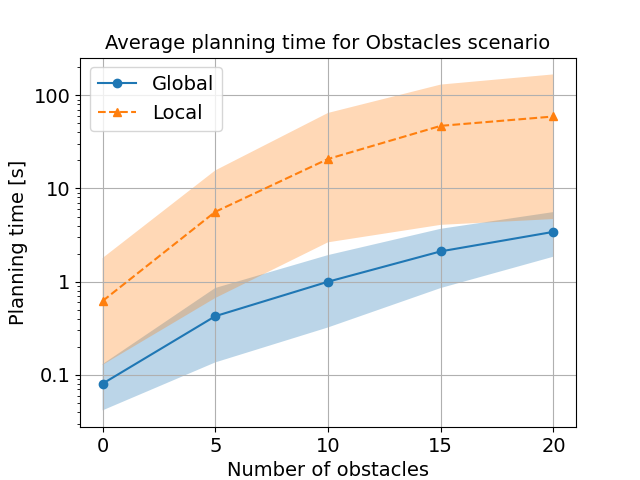}
        \caption{\small{Planning time vs. complexity}}
        \label{fig:results_obstacles_time}
    \end{subfigure}
    \caption{%
    \textbf{Known Environment Scenario}. 
    The robot (blue) must move a block (red) to the goal in the lower right corner of the map.
    Obstacles (grey) are randomly generated, ranging from $0$ to $20$ in number, to explore the effects of obstacle density on planning time (a).
    With the global reactive approach, even though each individual graph node expansion becomes more expensive as obstacle density increases (b), the relative decrease in node expansions (c) results in an overall decrease in planning time (d).
    Shaded areas denote 5th to 95th percentiles.
    \label{fig:obstacles}
    \vspace{-15pt}
    }
\end{figure*}

\subsection{Global Reactive Planning Approach}
\label{sec:global_reactive_planning}


The crucial advantage of the global reactive controller developed in~\cite{vasilopoulos_pavlakos_bowman_caporale_daniilidis_pappas_koditschek_2020} is that it guarantees successful navigation to any pose in its \emph{connected component of the freespace}, $\freespace$. 
Recall the function $\beta_{\freespace}(\robotposition)$ defined in Section~\ref{sec:reactive_layer} that maps each robot pose to the connected component of the freespace containing that pose. 
In the more general reactive planning setting, $\beta_{\freespace,o}(\configuration)$ depends on the configuration $\configuration$ of each object, and returns the connected component of the freespace of object $o$, conditioned on the pose of each other object. The reactive layer then defines, for any goal pose $\mathbf{o}^*$, a closed-loop controller with an attractor basin that includes the polygon $\beta_{\freespace,o}(\goalconfiguration)$. 
Formally, for the purposes of the deliberative layer, $\texttt{isfeasible}(\mathbf{o}, \mathbf{o}^*) = 1$, if $\mathbf{o} \in \beta_{\freespace,o}(\goalconfiguration)$.



Because the global reactive controller guarantees collision-free convergence to any target configuration in the connected component of the robot's (or mated robot--object's) freespace, the deliberative planning domain includes actions whose requirements and effects are defined by closed-loop controller attractor basins.
These are larger sets of the configuration space than the line-of-sight sets exposed by the local reactive planning approach. Any sequence of invocations of the reactive planner can be represented using these actions; because the actions can traverse long distances, this approach results in shorter plans, particularly in environments with unanticipated obstacles or complex geometric features.

We describe the global reactive planner as a C3 domain, and use the sampling-based planning algorithm described in Section~\ref{sec:combined_search} to search for a sequence of transitions between adjacent basins of attraction created by invocations of the reactive layer. To determine adjacency, we explicitly construct the polygonal connected components of the freespace containing each robot and object using the Boost Geometry library \cite{boost}.
Two polygonal components are adjacent if the distance between them is small enough to be traversed by a jumping or manipulation action.

\begin{figure*}[t]
  \vspace{2pt}
  \centering
  \begin{subfigure}[t]{0.2\textwidth}
    \centering
    \includegraphics[width=\textwidth]{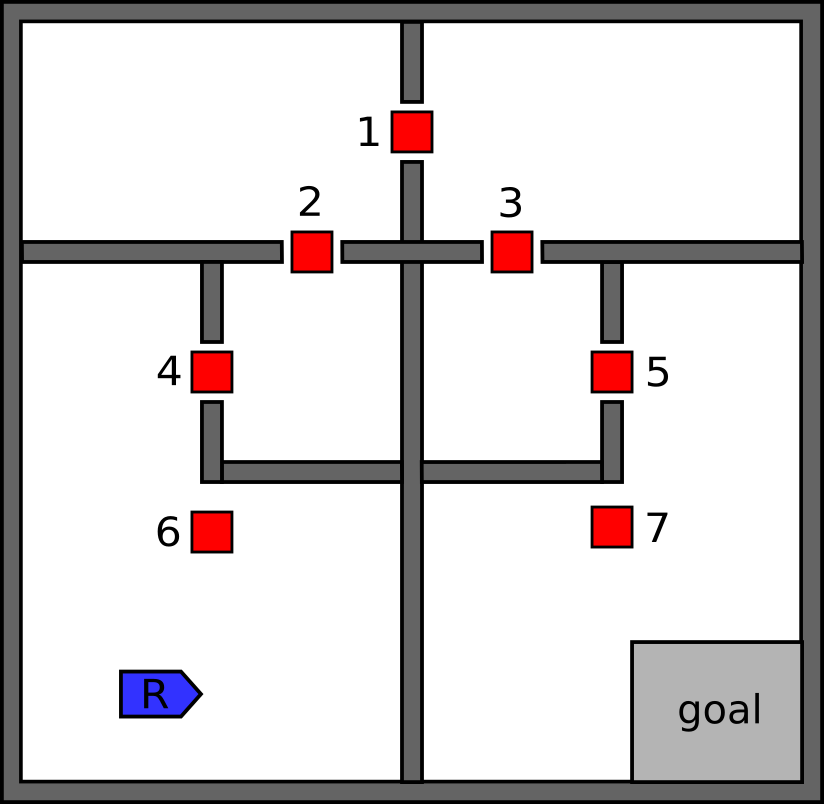}
    \caption{\small{Scenario overview}}
        \label{fig:scenario_doorways}
  \end{subfigure}
  \begin{subfigure}[t]{0.34\textwidth}
        \includegraphics[width=\textwidth]{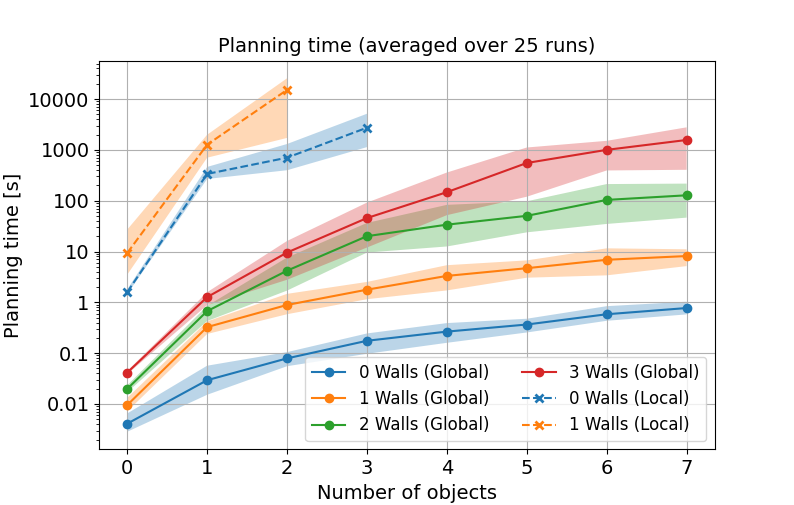}
        \caption{\small{Planning time vs. complexity}}
        \label{fig:results_doorways}
  \end{subfigure}
  \begin{subfigure}[t]{0.2\textwidth}
    \includegraphics[width=\textwidth]{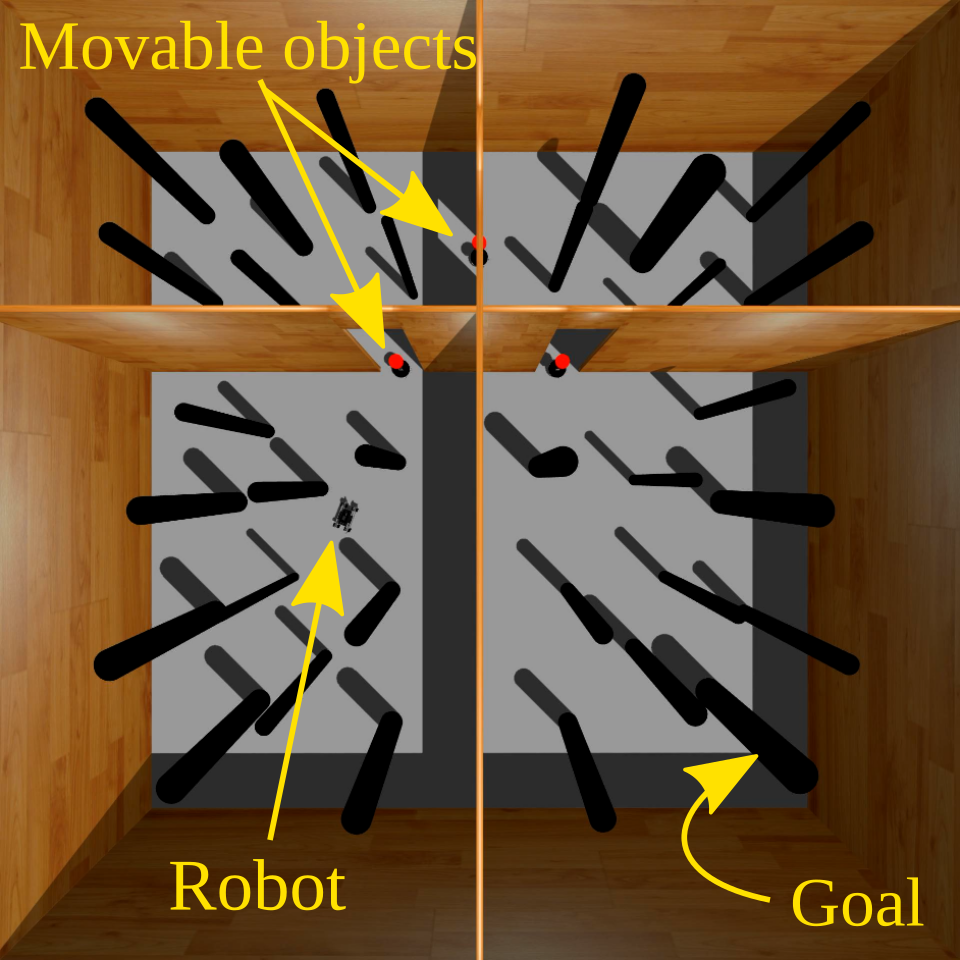}
    \caption{\small{Gazebo scenario}}
    \label{fig:gazebo_doorways_1}
  \end{subfigure}
  \quad
  \begin{subfigure}[t]{0.2\textwidth}
    \includegraphics[width=\textwidth]{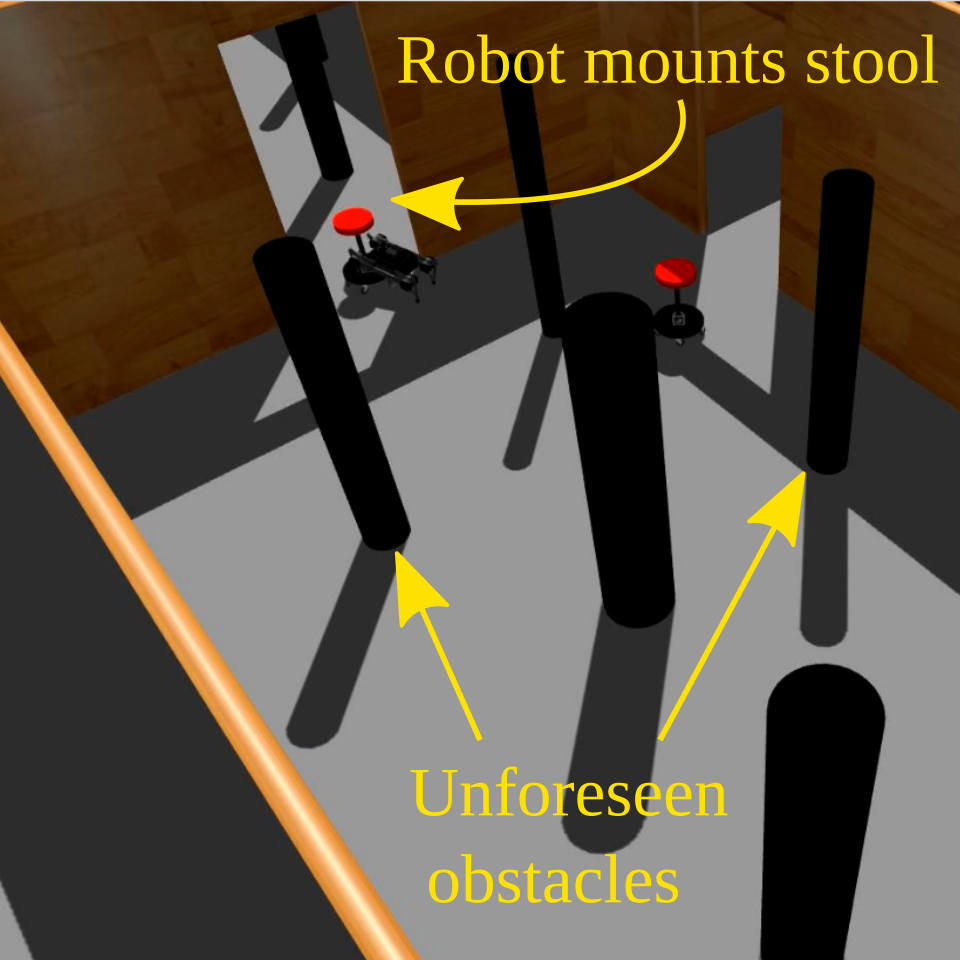}
    \caption{\small{Simulation details}}
    \label{fig:gazebo_doorways_2}
  \end{subfigure}
  \caption{%
    \textbf{Doorways Scenario}. 
    The robot (blue) must traverse an increasingly complex set of walls and push objects out of the doorways to reach the goal (a).
    Planning time increases with number of doorways and obstacles, which both add complexity to the problem.
    Additionally, the contract provided by the global reactive approach significantly reduces planning times for equivalent problems compared to when utilizing the local reactive approach. 
    Shaded areas denote 5th to 95th percentiles. (b). 
    Walls are added in the following order: first, the vertical wall containing doorway 1, then the horizontal wall with doorways 2 and 3, and finally the center walls containing doorways 4 and 5.
    Objects are randomly placed in locations 1--7, ensuring that first all existing doorways are blocked before placing extraneous objects in free space.
    We simulated this scenario in Gazebo (see the accompanying video submission), adding random cylindrical obstacles that were unknown to the deliberative planner (c), (d).
    By leveraging the proposed global reactive planning approach, the robot is able to navigate the environment and manipulate movable objects to reach a goal despite the unforeseen obstacles.
  }
  \label{fig:doorways}
  \vspace{-18pt}
\end{figure*}

%% file: 5-results.tex
\section{Numerical Experiments}
\label{sec:numerical_experiments}

In this Section, we present scenarios that describe common task specifications that can be solved using our system. 
We perform qualitative and quantitative analyses contrasting the performance of the deliberative planner with a local reactive and a global reactive planner by exploring the effect of environmental complexity on planning computation. 

\subsection{Known Environment Scenario}
\label{sec:static_obstacles_scenario}

In the scenario in Fig.~\mbox{\ref{fig:obstacles}}, the robot must move an object to a goal while navigating an increasingly dense set of randomly generated obstacles known by the deliberative layer.
Fig.~\ref{fig:results_obstacles_time} shows that planning times increase with the number of obstacles at a higher rate using the local reactive approach. 
Despite the added overhead of decomposing the environment into connected components when using the global reactive planning approach, more samples are needed to successfully solve these tasks using the local reactive system, where the deliberative planner is required to sample collision-free straight-line motion primitives at a lower level of abstraction.
Also, the success rate of planning with the local reactive approach decreases with environmental complexity.
For all generated worlds in which the global reactive approach found a plan, we planned $10$ times more with the local reactive approach. 
In the case of zero obstacles, average success rate was $100\%$, gradually decreasing to $66.5\%$ for $20$ obstacles.

\subsection{Doorways Scenario}
\label{sec:doorways_scenario}

The scenario in Fig. \ref{fig:doorways} explores how planning time scales with environmental complexity, both in terms of static obstacles and movable objects.
Since the complexity of walls is abstracted away with the global reactive layer, this scenario can be solved with less samples (and therefore in less time) than with the local approach, as shown in Fig.~\ref{fig:results_doorways}. Moreover, extraneous objects that do not block a doorway have a significantly lower impact on planning time, as seen in the inflection points for the ``3 Doorways'' and ``5 Doorways'' lines at 3 and 5 objects, respectively.
Planning time does increase when the number of walls (and thus doorways) increases while the number of objects is held fixed.
This is partially attributed to additional node expansions in sampling placements for objects, but also to more polygon decomposition and adjacency checks per individual node expansion.


\section{Simulation Experiments}
\label{sec:physical_experiments}

We demonstrate our proposed system architecture on a Ghost Minitaur~\cite{ghostminitaur} quadrupedal robot using the Gazebo simulator\footnote{
Video of these simulations is included in the video submission and online in \url{https://youtu.be/Ta5sVFkNnxo}. 
The files for simulating Minitaur in Gazebo can be found in \url{https://github.com/KodlabPenn/kodlab_gazebo} and a C++ implementation of the reactive layer is included in \url{https://github.com/KodlabPenn/semnav}.
}.
In our implementation, the gait layer abstracts the details of determining how to move the limbs or negotiate uneven terrain, freeing the reactive planner to determine how to achieve local goals for placements of the body and movable objects in the world. Specifically, we employ the steady-state behaviors {\it ``Walk''} and {\it ``Push-Walk''} from~\cite{Vasilopoulos_Topping_Vega-Brown_Roy_Koditschek_2018}, to either navigate the workspace or use the robot's front limbs as a virtual gripper when manipulating movable objects respectively. 
In addition, we use a set of four transitional behaviors: {\it ``Mount''}, {\it ``Dismount''}, {\it ``Jump-Up''} and {\it ``Jump-Across''}, adapted from~\cite{topping2019composition}, to mount and dismount objects, or jump on platforms or across gaps.


\subsection{Doorways Scenario}
\label{sec:sim_doorways_scenario}

We show the robot executing plans generated using our method on the \emph{Doorways} scenario of Section~\ref{sec:doorways_scenario}. 
Fig.~\ref{fig:gazebo_doorways_2} demonstrates that the global reactive approach allows the deliberative layer to find plans even in the presence of a complex space punctured by a large number of obstacles. Here, the robot has prior knowledge of all fixed walls, but no prior information on the location of the cylindrical obstacles; it must discover and avoid them using an onboard LIDAR.


\begin{figure*}[t]
  \vspace{8pt}
  \centering
  \begin{subfigure}[t]{0.22\textwidth}
    \centering
    \includegraphics[width=\textwidth]{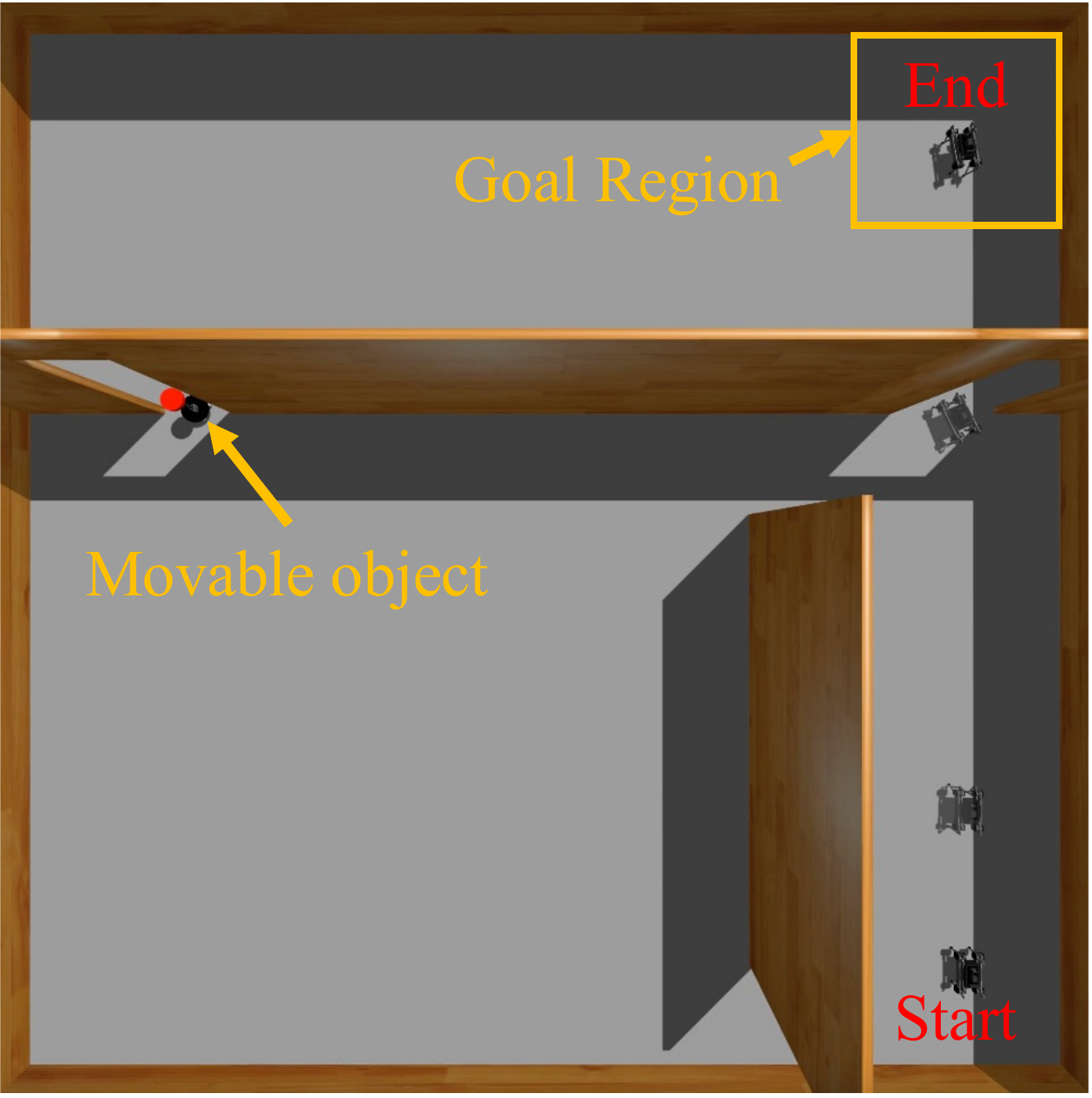}
    \caption{\small{No unforeseen obstacles}}
    \label{fig:replanning1}
  \end{subfigure}
  \quad
  \begin{subfigure}[t]{0.22\textwidth}
    \includegraphics[width=\textwidth]{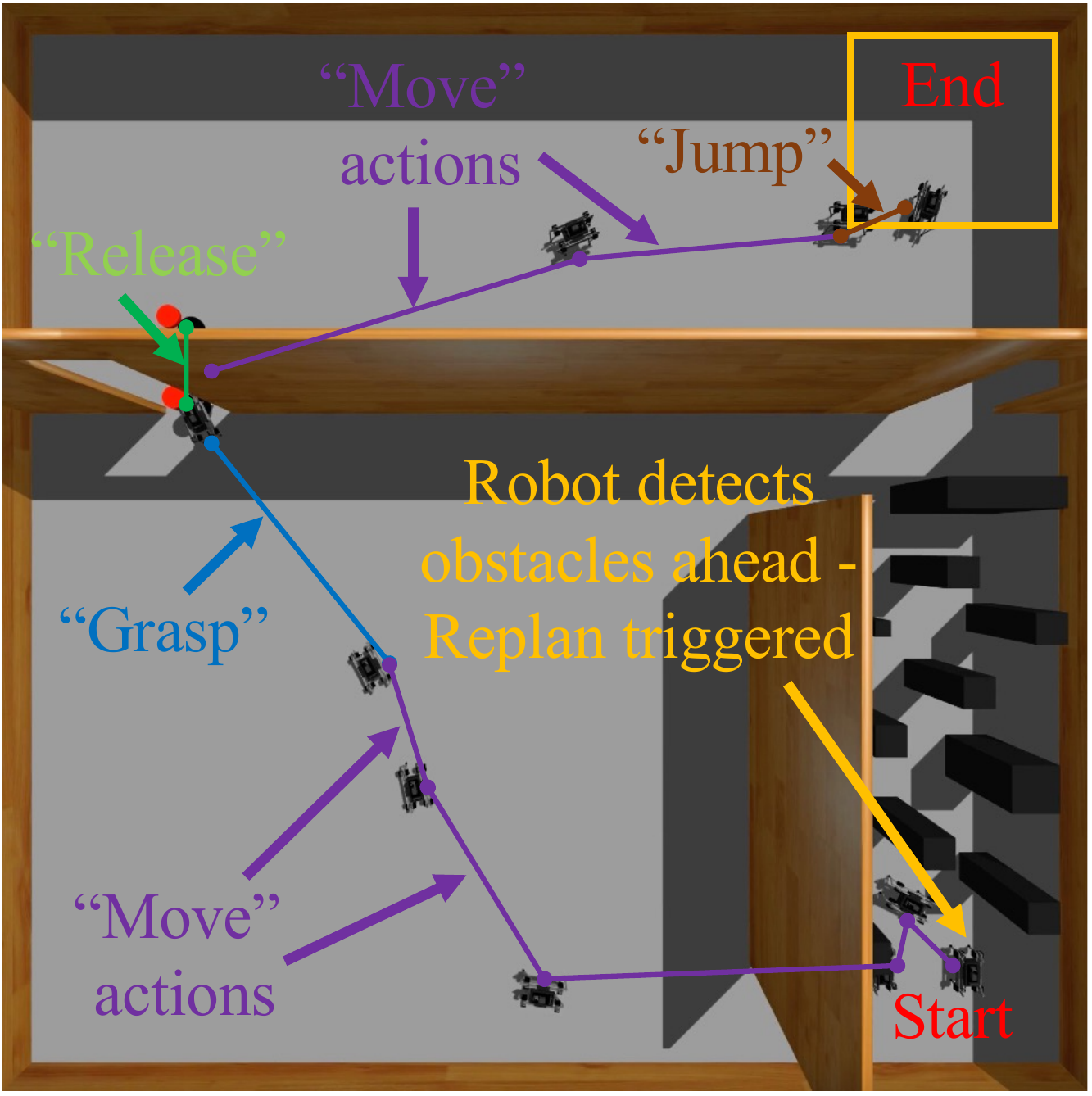}
    \caption{\small{Local Reactive Planning}}
    \label{fig:replanning2}
  \end{subfigure}
  \quad
  \begin{subfigure}[t]{0.22\textwidth}
    \includegraphics[width=\textwidth]{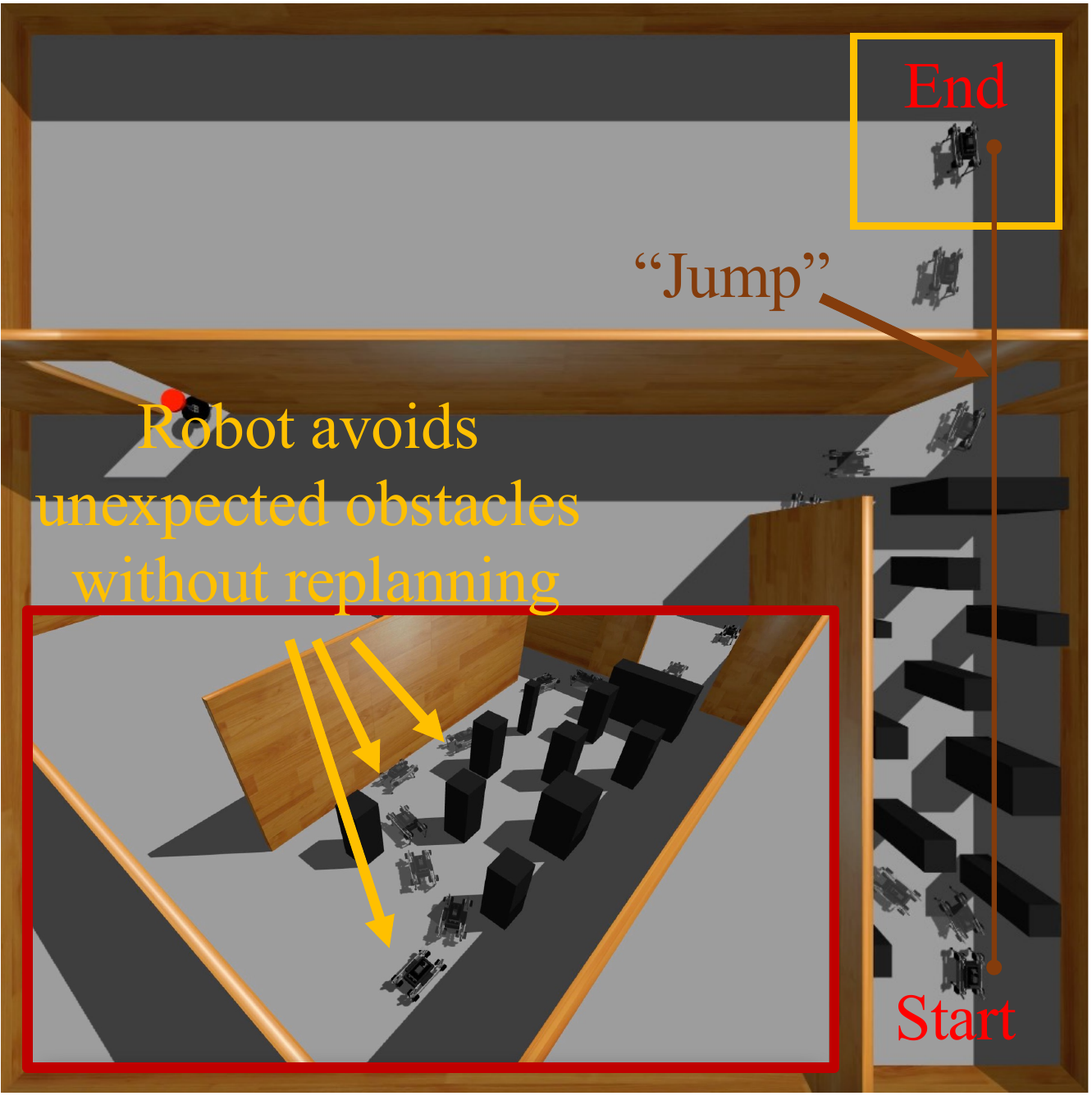}
    \caption{\small{Global Reactive Planning}}
    \label{fig:replanning3}
  \end{subfigure}
  \quad
  \begin{subfigure}[t]{0.22\textwidth}
    \includegraphics[width=\textwidth]{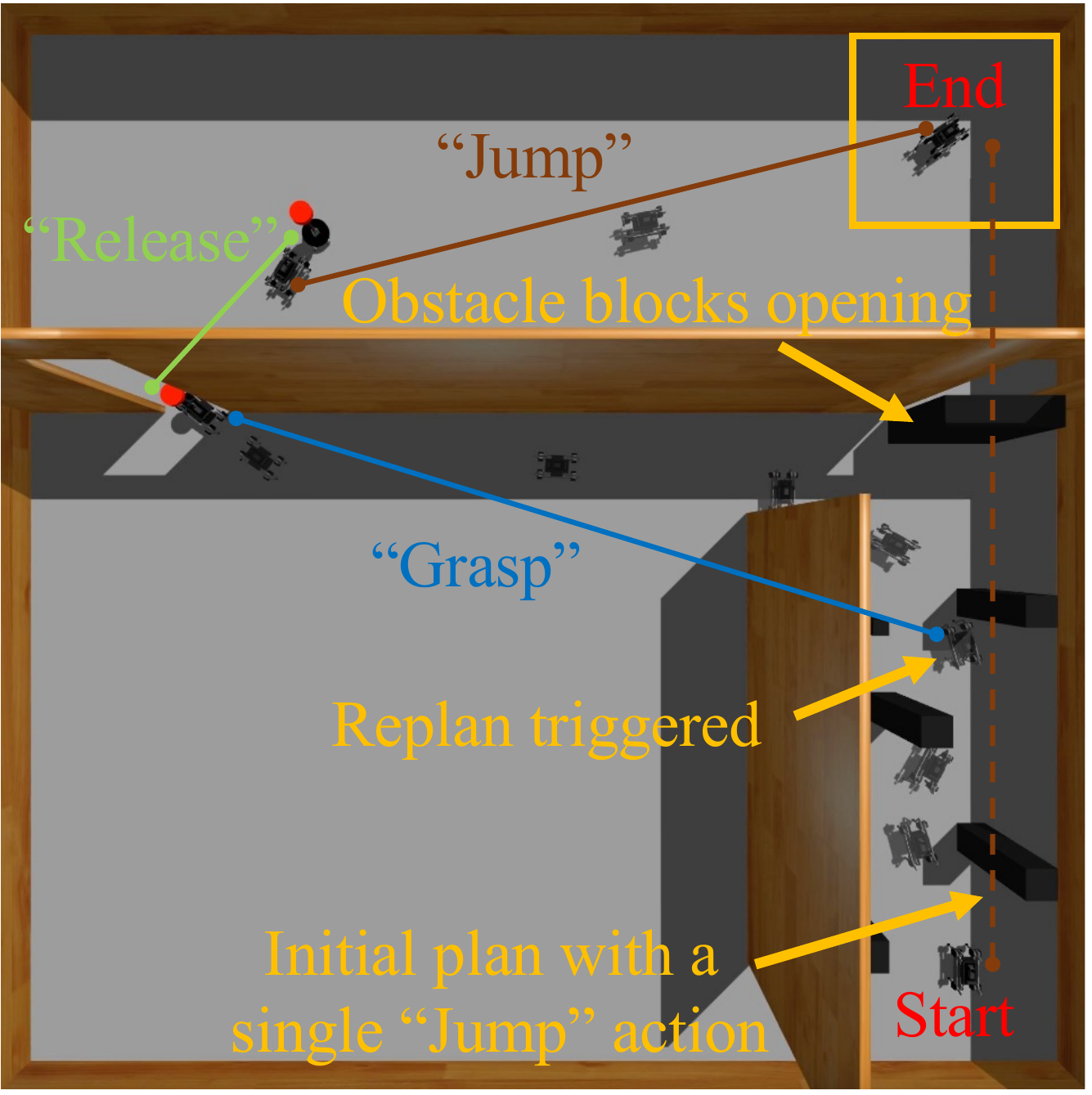}
    \caption{\small{Replanning needed}}
    \label{fig:replanning4}
  \end{subfigure}
  \caption{
    \textbf{Unknown Environment Scenario}. 
    Minitaur must move to a goal region at the top right of the environment.
    The lowest-cost solution involves navigating along the hallway on the right, which is successfully executed using both the local reactive and the global reactive approach (a).
    Random unanticipated obstacles (black) may appear in this hallway and are only detected and localized when the robot approaches them within a specified distance. 
    Using the local reactive approach, the robot quickly abandons the initially evaluated plan because some of the initial waypoints lie in obstacle space, and replans. 
    It then unnecessarily switches to a higher-cost plan, involving manipulating the movable object and navigating a longer path to the goal (b). 
    Using the global reactive approach, the robot either avoids all interior obstacles without changing its initially executed plan (c), or requests an alternative plan when detecting that the requirements of the contract are violated, i.e., the robot and the goal lie in different connected components of the freespace (d).
    }
  \label{fig:dynamic_obstacles}
  \vspace{-15pt}
\end{figure*}

\subsection{Unknown Environment Scenario}
\label{sec:dynamic_obstacles_scenario}

In this scenario, shown in Fig.~\ref{fig:dynamic_obstacles}, the robot must move from its starting pose to a specified goal pose.
In the absence of obstacles, the lowest-cost solution involves directly moving along a hallway to the goal.
As the robot discovers obstacles while moving down this hallway, it may be able to navigate around the obstacles and still reach the goal.
However, as the obstacle density increases, it may become difficult to plan around these obstacles, or the hallway may be blocked altogether.
In this case, there is a higher-cost alternative in which the robot can push a movable object near the top of the environment and navigate a longer path to the goal. For these simulations, we assume that the robot possesses a sensor of fixed range (set at 3 m), for localizing unexpected obstacles.

We qualitatively show that the global approach can handle unanticipated obstacles without triggering a full replan, unless a newly localized obstacle blocks the hallway, violating the contract between the layers.

%% file: 6-conclusion.tex
\section{Conclusion}
\label{sec:conclusions}


Our hierarchical planner exhibits the greatest gains in efficiency when finding long plans with a small number of actions; problems that require evaluating combinatorially many transitions remain an open challenge.
Our contract-based approach to action modeling could be combined with recent improvements in sampling strategies, search algorithms, and planning heuristics to increase the size and complexity of planning problems we can address. 
In addition, our approach could be applied to other classes of robotic platforms for which local controllers can be devised. 
Finally, we note that one shortcoming of our approach is the lack of a mechanism for the deliberative planner to correct the reactive planner if it makes a locally suboptimal decision. 
In principle, we could defer the selection of the level of abstraction at which to search for a plan to execution time, or even interleave searches at different levels of abstraction to capitalize on the relative strengths of each representation.